%% file: main.tex
\documentclass[9pt,twocolumn,twoside, pdflatex]{pnas-new}

\templatetype{pnasresearcharticle} 

\newcommand{\bX}{\mathbf{X}}
\newcommand{\bh}{\mathbf{h}}
\newcommand{\bz}{\mathbf{z}}
\newcommand{\MLP}{{\mathrm{MLP}}}
\newcommand{\CONCAT}{{\mathrm{CONCAT}}}
\newcommand{\NN}{{\mathrm{NN}}}
\newcommand{\GNN}{{\mathrm{GNN}}}
\newcommand{\FP}{{\mathrm{FP}}}
\newcommand{\IS}{{\mathrm{IS}}}

\title{Ensemble Spectral Prediction (ESP) model for metabolite annotation}

\author[a]{Xinmeng Li} 
\author[a]{Hao Zhu} 
\author[a]{Li-ping Liu} 
\author[a,b,1]{Soha Hassoun}
\affil[a]{Department of Computer Science, Tufts University, Medford 02155, USA}
\affil[b]{Department of Chemical and Biological Engineering, Tufts University, Medford 02155, USA}

\leadauthor{Li} 

\correspondingauthor{\textsuperscript{1}To whom correspondence should be addressed. E-mail: soha.hassoun@tufts.edu}


\begin{abstract}
A key challenge in metabolomics is annotating  measured spectra from a biological sample with chemical identities. Currently, only a small fraction of measurements can be assigned  identities. 
Two complementary computational approaches have emerged to address the annotation problem: mapping candidate molecules to spectra, and mapping  query spectra to  molecular candidates. In essence, the candidate molecule with the spectrum that best explains the query spectrum is recommended as the target molecule. Despite candidate ranking being fundamental in both approaches, no prior works utilized \textit{ rank learning} tasks in determining the target molecule.

We propose a novel machine learning model, Ensemble Spectral Prediction (ESP), for metabolite annotation.
ESP takes advantage of prior neural network-based annotation models that utilize multilayer perceptron (MLP) networks and Graph Neural Networks (GNNs). Based on the ranking results of the MLP and GNN-based models, ESP 
learns a weighting for the outputs of MLP and GNN spectral predictors to generate a spectral prediction for a query molecule.  Importantly,  training data is stratified by molecular formula to provide candidate sets during model training. 
Further, baseline MLP and GNN models are enhanced by considering peak dependencies through multi-head attention mechanism and multi-tasking on spectral topic distributions. ESP improves  average rank by 41\% and 30\% over the MLP and GNN baselines, respectively, demonstrating remarkable performance gain over state-of-the-art neural network approaches.  We  show that annotation performance, for ESP and other models, is a strong function of the number of molecules in the candidate set and their similarity to the target molecule. 

\end{abstract}

\dates{This manuscript was compiled on \today}
\doi{\url{www.pnas.org/cgi/doi/10.1073/pnas.XXXXXXXXXX}}

\begin{document}

\maketitle
\thispagestyle{firststyle}
\ifthenelse{\boolean{shortarticle}}{\ifthenelse{\boolean{singlecolumn}}{\abscontentformatted}{\abscontent}}{}

\input{1_introduction}

\input{fig_subproblems}

\input{fig_steps}

\input{2_results}

\input{3_materials}
\input{4_methods}

\input{5_discussion}

\showmatmethods{} 

\acknow{This research is supported byXXXX}

\showacknow{} 

\bibliography{reference}
\clearpage
\input{6_SI}
\end{document}

%% file: 1_introduction.tex
\dropcap{N}ontargeted tandem mass spectrometry is a powerful approach to characterize small molecules produced in cells, tissues, and other biological systems. Unlike genomics that specify the cell’s capabilities, metabolites are direct products of enzymatic reactions and provide an accurate functional readout of cellular state \citep{baker2011metabolomics, patti2012metabolomics}. So far, metabolomics studies have identified disease biomarkers, elucidated biochemical changes associated with drug responses, created opportunities for personalized medicine, and analyzed relationships between diet and health \citep{johnson2016metabolomics, chong2018metaboanalyst, jacob2019metabolomics, kitano2002computational}. Importantly, the ability to collect thousands of measurements on the sample-under-study promises to broadly profile the metabolome and revolutionize phenotyping and advancing biological discovery.

A key challenge in metabolomics is the  "annotation" problem, where measured spectra  are assigned chemical identities. Currently, only a small fraction of  measured spectra is annotated \citep{da2015illuminating}. As spectral libraries are limited and experimental exploration is costly and time-consuming, computational approaches have emerged as an effective complementary alternative \citep{liebal2020machine}. Translating between  molecular and spectral views have given rise to spectrum-to-molecule  and molecule-to-spectrum approaches. In the first approach, a query spectra is mapped to a molecular structure or properties. CSI:FingerID \citep{duhrkop2015searching}  predicts properties molecular fingerprints for a set of candidate molecules, and then ranks the query spectra against the predictions. MassGenie \citep{shrivastava2021massgenie} casts the spectrum-to-molecule problem as a  translation problem from binned mass spectral peaks to a SMILES string using the  deep-learning transformer model \citep{vaswani2017attention}. MassGenie then uses a variational autoencoder to generate de novo candidate molecules that are 'close' in the chemical space.
In the second approach, spectra is predicted for candidate molecules that match to the chemical formula of the measured spectra. The candidate spectra are then ranked against the query spectra based on spectral similarity.  The candidate molecule with the highest scoring spectra is then assigned as the identity of the query spectra. Example approaches include MetFrag \citep{ruttkies2016metfrag}, CFM-ID \citep{allen2014cfm}, and our recent Graph Neural Network-based (GNNs) method \citep{zhu2020using}.
None of the prior works, however, regardless of approach,  utilize  \textit{learning} tasks to  distinguish the  ranking of the candidates against the query. 

We provide in this paper a conceptual framework and an accompanying deep-learning  approach for solving the molecule-to-spectrum problem. The problem is conceptually examined as three subproblems (Fig. \ref{fig:subproblems}): 
 1) representation learning of candidate molecules, 
 2) mapping candidates to their corresponding spectra, and 
 3) learning to rank the predictions of the candidate molecules against the measured spectra.  
 Learning molecular representations, as opposed to fixed fingerprints,  allows for flexible representations that may lead to improved top ranked candidate predictions. 
 Mapping from candidates to spectra is a translation task. Neural network models excel at learning such tasks. 
 Learning to rank  candidates however is a  challenging  task as it implies, at an initial glance, the need for  molecular candidate sets and their spectra as training data.  Importantly, this conceptual framework can also applied to the spectrum-to-molecule problem (Supplementary Fig. S1), where the ranking can be learned on preselected  molecular candidates or on de novo candidates. 

To implement this framework, we present in this paper a novel molecule-to-spectrum neural-network based ensemble model, referred to as Ensemble Spectral Prediction (ESP). The ensemble model utilizes both MLP and GNN-based spectral predictions  as each can outperform the other for differing molecules. Spectral predictions using MLP and GNN approaches are enhanced by using multi-head attention mechanism to capture dependencies among spectral peaks. Multi-tasking on additional data (spectral topic labels obtained using LDA (Latent Dirichlet Allocation \citep{blei2003latent, van2016topic}) is used to further improve MLP and GNN spectral predictions. Importantly, we circumvent solving the challenging problem of retrieving candidate molecules (and their spectra) for our training data. Instead, we partition the molecules in the training set based on their chemical formulas and \textit{learn}, based on \textit{ranking} results, how to combine the MLP and GNN-based spectral predictions to improve  spectral prediction. 
Measuring performance using average rank and rank@k on rank learning through the ESP model shows remarkable gain over existing neural-network approaches.

%% file: fig_subproblems.tex
\begin{figure*}
\centering
\includegraphics[width=\linewidth]{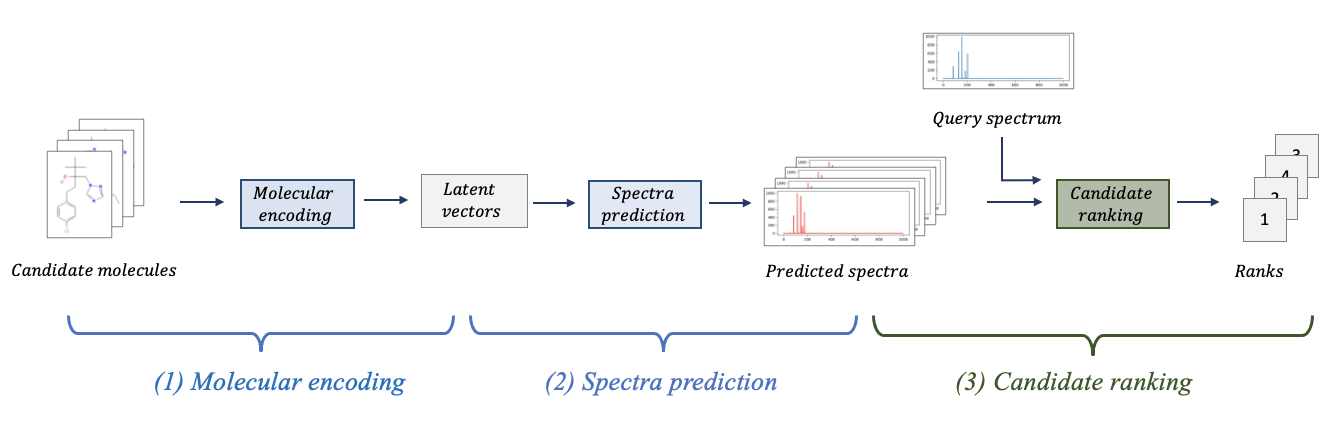}
\caption{A conceptual framework for solving the three subproblems involved in the molecule-to-spectrum  annotation problem: (1) representation learning of molecules, (2) spectra prediction from latent representation vectors, and (3) 
 ranking of candidate spectra against query spectrum using spectral similarity.}
\label{fig:subproblems}
\end{figure*}

%% file: fig_steps.tex
\begin{figure*}
\centering
\includegraphics[width=\linewidth]{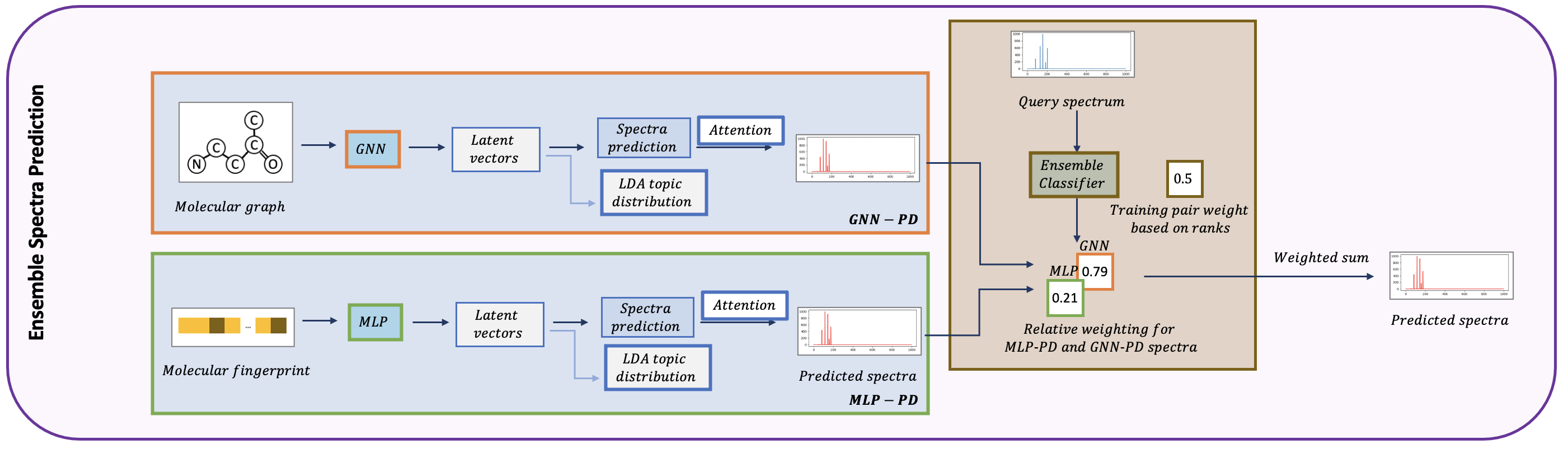}
 \caption{\textbf{The Ensemble Spectra Prediction (ESP) model has two phases. } \textbf{Phase 1.} 
 Molecular encoding using GNN and MLP followed by spectra prediction enhancement using  multi-head attention to capture spectral dependencies and using multi-tasking on predicting LDA topic distribution. 
 \textbf{Phase 2.} Training ensemble scoring model, we compare the rank of spectra prediction from GNN and MLP encoding on candidate ranking problem for query spectra in the training set to predict a score on weighing the two models. Note that ESP focus on solving the three subproblems of metabolite annotation: GNN and MLP models in phase 1 focus on (1) molecular encoding; attention and multi-tasking on LDA topic distribution prediction in phase 1 focus on (2) spectra encoding; ensemble scoring based on ranks of target molecule from two models in phase 2 focus on (3) candidate ranking.}


\label{fig:steps}
\end{figure*}

%% file: 2_results.tex
\section*{Results}
\subsection*{ESP overview}
The ESP model improves on current MLP and GNN models by capturing spectral peak dependencies and by  combining  spectral predictions of the GNN and MLP-based models to achieve the highest candidate ranking. The ESP model is trained in two phases (Fig. \ref{fig:steps}). In phase 1,  GNN and MLP-based models are trained to predict the spectra. In phase 2, the spectra prediction models are evaluated on candidate ranking, and the ranking results are used to train an ensemble classifier to judiciously weigh the MLP and GNN spectra predictions. Phase 1 thus addresses the first two subproblems and phase 2 addresses the third subproblem in Fig.~\ref{fig:subproblems}.

\begin{itemize}
    \item \textbf{Phase 1. Molecular encoding and spectra prediction.}
   MLP- and GNN-based models are trained in end-to-end fashion to encode the molecules and predict the spectra. As input, the MLP model utilizes the ECFP molecular fingerprint while the GNN model utilizes the molecular graph. Both models predict the intensity of the spectra binned at 1Da intervals. 
    To improve over prior MLP \citep{wei2019rapid} and GNN  \citep{zhu2020using} spectra prediction models, we  test various GNN models \citep{xu2018powerful, velivckovic2017graph,schlichtkrull2018modeling,lei2017deriving} and select the best performing model, Graph Isomorphism Network with Edge features (GINE)\citep{hu2019strategies}.  
   Bidirectional prediction, originally proposed for MLP models \citep{wei2019rapid}, is added to the GNN models to aid  in predicting intensities of larger fragments.
   

    Spectra prediction accuracy is enhanced by incorporating peak dependencies using attention mechanism and spectral topic distributions (motifs).
    Multi-head attention \citep{vaswani2017attention} is applied to capture pairwise peak dependencies. Through multi-task learning, ESP predicts spectral motifs that are learned on our dataset through topic modeling \citep{blei2003latent, wallach2006topic}. As in prior works \citep{van2016topic},  peaks are assumed words, and spectra are assumed documents. Each spectral topic is modeled as a probability distribution over a vocabulary of peaks (mass-to-charge (m/z) values and their intensities), while each spectra is modeled as a probability distribution over topics. The latter is referred to as a spectral motif.
    The MLP and GNN models combined with peak dependencies (PD) are referred to as MLP-PD and GNN-PD, respectively.
    Further, experimental instrument settings are used in  MLP and GNN-based models, thus explicitly accounting for collision energies and instrument types, and allowing for spectra predicting and candidate ranking under a wide range of instrument settings.  
    
 %

    \item \textbf{Phase 2. Ensemble model training based on candidate ranking.} 
We train an ensemble model that judiciously combines spectra predicted by the trained GNN-PD and MLP-PD models. Because we aim to solve the candidate ranking problem, the ensemble model is trained to leverage the ranking capabilities of the two models. 
As the ranking task requires candidate molecules/spectra, our training dataset is  stratified by molecular formula, thus forming molecular candidate lists with known spectra. Each paired spectrum/molecule therefore utilizes spectra/molecules within its strata as candidates. As candidates from molecular databases typically do not have known spectra (if they did, we would have used them for training!), this stratification strategy allows us a rich molecular candidate dataset with known spectra.

The MLP-PD and GNN-PD are used to rank the candidates.  Ranking performance is  reported using spectral similarity and the rank of the target molecule within the candidate set. Each training example is thus assigned a ranking under each model. Additionally each example is assigned a label to indicate which of the two models, GNN-PD or MLP-PD, outperforms the other in terms of ranking. Based on the ranking positions and their differences, a score is assigned to each example to indicate its importance  in distinguishing between the ranking performance of the two models. 
The importance score is computed using symmetric mean absolute percentage error (SMAPE) and provides higher weights for lower ranks while favoring training examples with better rank performances. 
Using the importance weights, a classifier, referred to as the \textit{ensemble classifier}, is then trained on the GNN/MLP labels. During test, the predicted label probabilities for a candidate spectra are used to weight the MLP-PD and GNN-PD predicted spectra. 

\end{itemize}

The ESP model is trained on the NIST-20 LC-MS/MS spectra data under positive mode with precursor type [M+H]+, with different collision energy levels \citep{phinney2013development}. Spectra is binned into 1000 bins, each covering a 1 Da range of mass-to-charge (m/z) ratios. To avoid over representing  the molecules with numerous spectra under different collision energy levels,  a maximum of  5 representative spectra is randomly selected per molecule with replacement. 
We utilized 89,405 spectra, involving 17,881 molecules. 
To ensure molecules in the test set are not in the training and validation sets, we split the data on molecules. The data was split randomly for training, validation, and test sets at the ratio of 8:1:1.
The test set had 12,375 spectra that correspond to 2,475 molecules.

During test, molecular candidates are retrieved from PubChem \citep{kim2016pubchem} for each test molecule, where  candidates have the same molecular formula as the target molecule. To provide consistent comparisons of our results across different models, candidate sets are sampled  to allow for a fixed average size of  molecules per candidate set. The default average size is 100 molecules, unless noted otherwise.  ESP is then applied to measured spectra in the test set and its corresponding candidate set. The ESP model  first predicts the spectra under the MLP-PD and GNN-PD models. Then, it  combines the spectra based on the probabilities predicted by ensemble classifier on the measured spectra. This last step  determines the relative weighting of the MLP-PD and GNN-PD spectral predictions. The ESP model then recommends the molecular candidate whose predicted spectra has the highest spectral similarity to the measured spectra.  We report our results by comparing against the baseline MLP \citep{wei2019rapid} and GNN \citep{zhu2020using} models.  On par comparisons with other tools \citep{verdegem2016improved,duhrkop2015searching, duhrkop2019sirius,wang2021cfm} provides limited insights into the performance of the underlying models due to the inaccessibility to  models trained on our same exact training/test splits. 
Further, the data split during training/testing can be detrimental to performance. In prior work \citep{duhrkop2015searching}, it was demonstrated that splitting training and test sets based on spectra (instead of molecules) can lead to significantly higher performance due to the same molecule appearing in both training and test sets. Here, we demonstrate  data split issues when training on well known molecules and testing on lesser known molecules.  In addition, we show  that ranking results are a strong function of the size and composition of the candidate sets.


\subsection*{ESP outperforms MLP and GNN-based models}
\input{table1}

To compare the  performance of the ESP, MLP and GNN-based models, we report the  average rank of the target molecule within the candidate set, and the average rank@k with $k={1, 3,10}$. The rank@k metric reflects the likelihood of identifying the target molecule among the \textit{k}-top ranked candidates, and the rank@k results are averaged across the test set. On all metrics (Table \ref{Tab:02}), 
ESP outperforms the baseline GNN and MLP models, as well as the  two models augmented the peak dependency analysis, MLP-PD, and GNN-DP. 
The average rank performance of 4.621 is a significant improvement over the baseline MLP and GNN  models. 
Augmenting the MLP and GNN baseline models with peak dependencies improves the average rank for both models.
However, MLP-PD improves  by 15\% over the MLP baseline, while GNN-PD improves by 8\%. 
On average,  ESP provides the  correct chemical identification of a query spectra when examining the top 5 candidates in  a  100-molecule average size candidate set. MLP and GNN-based models require a minimum of 6 candidates. 
ESP improves more on the average rank than on the rank@1, rank@3, and rank@10.

\input{fig_lossRank}

Preliminary analysis of the GNN and MLP model performance on test sets spurred the idea of combining the two models. 
For our test set, GNN ranks the target molecule higher than MLP for 24\% of the target molecules, while MLP ranks higher on 25\% of target molecules. 
GNN and MLP tie for 51\% of the cases. 
To further understand performance differences, we plot the correlation between  rank differences of the MLP-PD and GNN-PD models 
and their respective  spectral loss  between measured and predicted spectra for the test set (Fig. \ref{fig:lossRank}). 
In most cases, the spectral prediction loss and ranking differences are in agreement for the two models (blue points in the upper right and lower left quadrants). It is tempting to conclude that spectral prediction loss is a good proxy for the desired ranking tasks. However,  the correlation between the spectral loss and the ranking  is weak ($R^2$ score is 0.579). Therefore,  the spectral prediction loss may not strongly predict candidate ranking.
As the goal of annotation is to achieve the best possible ranking performance, combining the MLP and GNN spectral outputs based on the ranking results (and not on the spectral loss differences) is a natural next step, and allows benefiting from both GNN and MLP-based  spectra prediction models.


\subsection*{Training on MLP and GNN-based models ranking benefits ESP}
ESP is trained on the rankings obtained from  MLP-PD and GNN-PD. Further, each example is weighted by SAMPE. We first assess 
the benefits of using ranking instead of spectral loss. 
We develop a model, ESP-spectra\_loss (ESP-SL), where the ensemble classifier is trained using importance weights in proportion to the spectral loss differences (not rank differences). 
Specifically, the ensemble classifier is trained on the GNN/MLP labels now generated based on the smaller of the two predicted spectra losses.  ESP-SL achieves a 
6.426 average rank, thus performing somewhere between the MLP-PD and GNN-PD models.

Next, to evaluate the benefit of weighting each sample via SMAPE instead of uniform weighting, we develop a model ESP-rank\_uniform (ESP-RU), 
where
the ensemble classifier is trained on the GNN/MLP labels  generated based on rank results, but each training example is weighted uniformly.
ESP-RU underperforms ESP, where the former incurs a 6\% drop in average rank. However, the drops in  rank@k  are not as significant as the average rank. The discrepancy of the changes in average rank vs rank@k indicates the presence of "difficult" to rank molecules. The average rank vs rank@k performance differences are  consistent for all models (Table \ref{Tab:02}, and Supplementary Fig. S2).

\subsection*{Candidate quantity and quality dictate ranking performance}
\input{fig_alter}


We utilized a fixed average candidate set size of 100 molecules to facilitate our model evaluation efforts. However, 
the presence of difficult-to-rank molecules even within a fixed average candidate set size led us to analyze and explore candidate molecular sets retrieved from PubChem. Candidate set sizes varied tremendously, where some chemical formula retrieval yielded over 10,000 molecular candidates (Supplementary Fig. S3A). Further, pairwise (MACCS) fingerprint  similarity analysis between each test molecule and its candidates revealed a wide range of similarities (Supplementary Fig. S3B).  Our hypothesis is that larger candidate sets and those with many similar candidates to the test molecule are more challenging than their counterparts.  We explore this hypothesis by
varying the size and diversity of the candidate sets for molecules in our test sets.

The ESP model was used to predict and rank spectra for the same test target molecules but with different number of candidates molecules (Fig. \ref{fig:alter}A). The  candidate sets with differing sizes were generated 
by randomly sampling from the candidates retrieved from PubChem.
We vary the candidates size to 50, 100, 250 and 1000 molecules and report on  rank@k, for $k=1$ to $k=20$. Clearly,  smaller candidate sizes yield better candidate ranking, concluding that the size of the candidate set has a significant impact  on candidate ranking performance.  The same trend holds for the MLP-PD and GNN-PD models (Supplementary Fig. S4 A, B and C).

To assess the difficulty in ranking due  to the  similarity of the candidate molecules to the target molecule, two 100-molecule candidate sets  generated for each target test molecule: those that are most and least similar to the test molecules.  Molecular similarity is  calculated using pairwise MACCS fingerprint between  candidate  and target molecules. ESP performs best on the least similar candidates (Fig. \ref{fig:alter}B) and shows higher rank performance until rank 5, after which the model has comparable rank@k performance with randomly selected candidate sets. 
As expected, the model is more challenged with sets of candidates most similar to the target molecule than when using a randomly selected set of candidates. The rank@k performance decreases for all shown ranks. Importantly, a high similarity yet smaller candidate set (100 molecules) results in poorer performance when compared to a large (1000 molecules) randomly selected candidate set, up to the rank of 18.  The performance of MLP-PD and GNN-PD follow a similar trend to ESP (Supplementary Fig. S5 A and B). Further, MLP-PD outperforms ESP for rank 1-3 on the most similar dataset, highlighting the value of domain expertise in the fixed fingerprints and gaps in GNN representation learning.

\subsection*{Realistic data split}
Another source of difficulty in ranking candidates is the lack of similar molecules or spectra in the training dataset. 
While some molecules are popular in databases due to their known biological or chemical significance, such as their role in primary metabolism or their environmental relevance, there is limited documentation for many metabolic products.
We evaluate how the model performs under such a "realistic split" scenario \citep{martin2017profile}, 
where the model is trained on well annotated molecules  but tested on less popular molecules. 
Molecules are clustered based on  MACCS molecular fingerprint similarity.  UPGMA (the unweighted pair group method with arithmetic mean) \citep{sokal1958statistical} clustering method is applied on the first two dimension of t-SNE for the molecular fingerprint space  (Supplementary Fig. S6A). This clustering method results in 50 generated spectra.  The larger 29 clusters where used for training, while the smaller 21 clusters were used for testing. This specific split was selected to ensure models trained on realistic split are exposed to  similar amount of training data as used in random split. 
 ESP ranking under the realistic split drops when compared to the performance on a random split (Fig. \ref{fig:alter} C), but the gap under the two splits narrows at higher ranks. 
 This drop in performance is consistent for the the MLP-PD and GNN-PD models (Supplementary Fig. S6B). The performance of the GNN-PD is a close match to that of ESP, highlighting again the importance of learned representations. 
 
\subsection*{Full positive mode model performance}
Provided with the appropriate data, the ESP model can be trained to predict spectra under various  instrument settings, including 
precursor types and collision energies. To evaluate ESP's performance on  multiple instrument setting, we train and replicate the experiments on all positive mode spectra from the NIST20 dataset with various precursor types such as [M+H]+, [M+H-H2O]+, [M+H-2H2O]+, [M+H-NH3]+, and [M+Na]+.
The relative performance of the ESP models on all positive mode spectra is consistent with the performance of ESP on only [M+H]+ precursor types with some degradation (Fig. \ref{fig:alter}D).  The performance  trend is similar for the GNN-PD and MLP-PD models trained under multiple instrument settings (Supplementary Fig. S6C and Supplementary Table S1).

%% file: table1.tex
\begin{table}[!t]

\caption{Metabolite annotation evaluation on [M+H] precursor mode and average 100-molecule candidate set size \label{Tab:02}} {
\begin{tabular} {ccccc}
\toprule 
& Average rank & Rank@1 & Rank@3 & Rank@10 \\

\cmidrule(lr){2-2} \cmidrule(lr){3-5} 

& \multicolumn{1}{c}{\textcolor{blue}{The lower the better}} & \multicolumn{3}{c}{\textcolor{blue}{The higher the better}}\\

\midrule
MLP		&	7.768	&	0.530	&	0.716		&	0.859\\
GNN     &	6.629	&	0.588	&	0.778		&	0.894\\

\cmidrule(lr){1-5} 

MLP-PD	    &	6.618	&	0.620	&	0.784		&	0.894\\
GNN-PD	    &	5.981	&	0.624	&	0.799		&	0.906\\

\cmidrule(lr){1-5} 

ESP-SL	&	6.426	&	0.621	&	0.784		&	0.893\\

ESP-RU	&	4.903	&	0.631	&	0.818		& 0.916\\

\cmidrule(lr){1-5}

\textcolor{red}{\textbf{ESP}} 	&	\textcolor{red}{\textbf{4.621}}	&\textcolor{red}{\textbf{0.666}}	&	\textcolor{red}{\textbf{0.824}} 	&	\textcolor{red}{\textbf{0.921}}\\
\bottomrule
\end{tabular}}{}\
\end{table}

%% file: fig_lossRank.tex
\begin{figure}
\centering
\includegraphics[width=.8\linewidth]{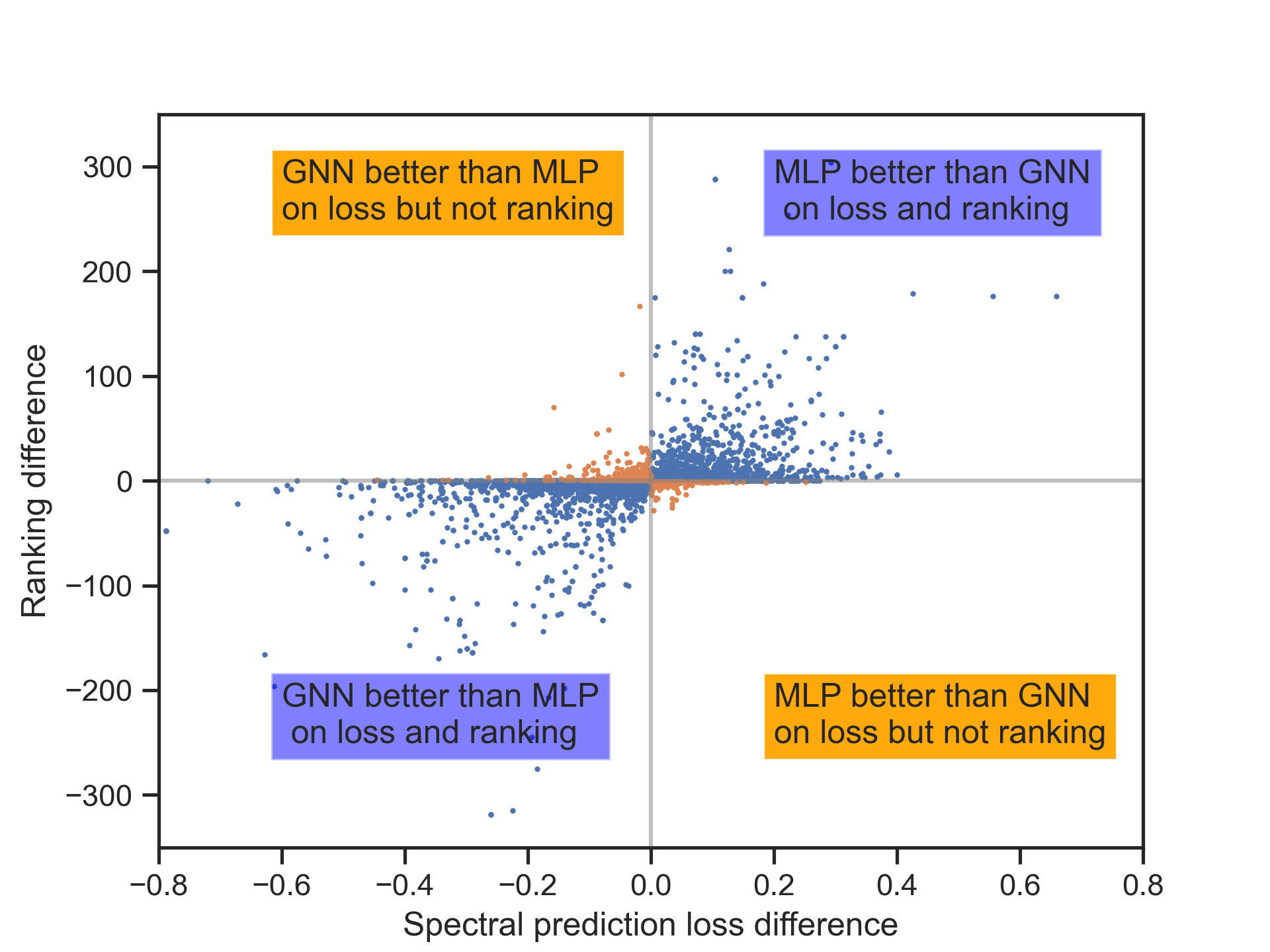}
\caption{Spectral loss difference (x-axis) versus rank difference (y-axis) between MLP-PD and GNN-PD results for test set molecules. 
Spectral prediction loss is computed using the negative cosine similarity between measured and predicted spectra.
Candidate ranking performance is the rank of target molecule per each model.
The rank and loss difference metrics are in agreement in some cases (blue points in two quadrants), but there are cases of disagreement (orange points). Importantly, the correlation of the two metrics is weak (R$^2$ score is 0.579). }
\label{fig:lossRank}
\end{figure}

%% file: fig_alter.tex
\begin{figure*}
\centering
\includegraphics[width=.95\linewidth]{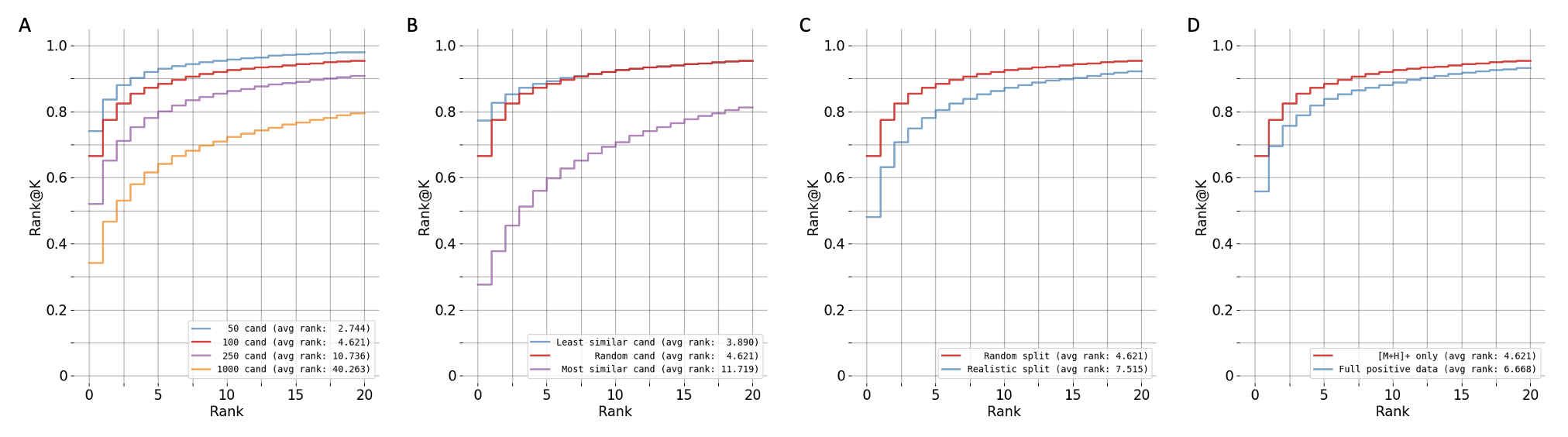}
\caption{Candidate ranking performances (rank@k) for various settings. Results are reported for a candidate set of 100, unless otherwise noted. A) Different number of candidates (cand) (50, 100, 250, 1000). B) Least and most similar 100 candidates. C) Random data split versus realistic split assuming 100 candidates. D) ESP performance on full positive dataset vs ESP performance on precursor [M+H]+ only.}
\label{fig:alter}
\end{figure*}

%% file: 4_methods.tex
\section*{Methods}

\subsection*{ MLP- and GNN-based Molecular encoding} 
The MLP-based models encode the ECFP molecular fingerprint, while the GNN-based models encode a molecular graph.
The models therefore learn a latent molecular representation vector,  $\bz$:
\begin{align}
    \bz = \bz_\MLP \quad \mathrm{OR} \quad \bz_\GNN
\end{align}
where $\bz_\MLP$ and $\bz_\GNN$ will be later defined in \eqref{eq_MLP} and \eqref{eq_GNN}, respectively.

We first describe the $\MLP$ model.
 In addition to the fingerprint, the MLP model encodes  instrument settings, which  specifies a one-hot encoding vector of the precursor type, if targeting more than one precursor type,  and a value corresponding to the normalized collision energy. The fingerprint and the instrument settings are encoded by a single-layer fully connected neural network, $\mathrm{NN(\cdot)}$, to predict the vector $\bz_\MLP$:
\begin{align}
        \bz_{\MLP} &= \MLP(\bX)
    \label{eq_MLP}
\end{align}
with $\bX$ includes both fingerprint $\bX_{\FP}$ and instrument setting $\bX_{\IS}$ as features,
\begin{align}
        \MLP(\bX) = \NN\left(\CONCAT\left({\NN(\bX_{\FP}}), ~ \NN(\bX_{\IS})\right)\right)
\end{align}

The GNN model encodes a molecular graph, $G=(V,E)$,  where graph nodes $v\in V$ correspond to atoms, and graph edges $(u,v)\in E$ correspond to bonds. We augment node features with the instrument settings. The GNN latent representation vector $\bz_{\GNN}$ is defined as:
\begin{gather}
    \bz_\GNN = \GNN(G)
    \label{eq_GNN}
\end{gather}
We use Graph Isomorphism Network with Edge features (GINE) \citep{hu2019strategies} to encode $G$. Besides GINE, we  explored a number of GNNs, including Graph Isomorphism Network (GIN) \citep{xu2018powerful}, Graph Attention Networks (GAT) \citep{velivckovic2017graph}, Relational Graph Convolutional Networks (R-GCN) \citep{schlichtkrull2018modeling}, and the Weisfeiler-Lehman network (WLN) \cite{lei2017deriving}, which was utilized in our prior work \citep{zhu2020using}. Our experiments showed that GINE outperformed  other models, potentially due to GIN's provably maximum discriminative power among GNNs \citep{xu2018powerful}.

We describe the  GINE model with $K$ layers. Node features contain atom information, including atom type and atom mass $\bX_{\mathrm{ATOM}}$, and instrument setting $\bX_\IS$; they are encoded by a single-layer fully connected neural network. The node representations are initialized as $\bh_v^{0}$:
\begin{align}
    \bh_v^0 =  \NN\left(\CONCAT\left({\NN(\bX_{\mathrm{ATOM}}}), ~ \NN(\bX_{\IS})\right)\right)
\end{align} 

Edge features represent bond types.  Edge representation at the $k$-th layer $\bh_e^k$ is computed from the encoded bond type $\bX_e$ as $\bh_e^k = \NN(\bX_e)$. 
Assuming $\bh_v^k$ is the representation of node $v$, \eqref{eq_GNN} is defined as:
\begin{gather}
    \mathbf{m}_v^k = \sum_{u: \left(u, v\right) \in E} \bh_u^{k-1} + \sum_{e=(u,v) \in E} \bh_{e}^{k-1} \\
    \bh_v^k = \mathrm{ReLU}\left( \NN^k (\mathbf{m}_v^k) \right) \\
     \bz_\GNN = \mathrm{READOUT} (\bh_v^{K}) 
\end{gather}
where $\mathbf{m}_v^k$ is the message for node $v \in V$, updated from its neighboring nodes ($u: \left(u, v\right) \in E$) and connected edges ($e=(u,v) \in E$). The $\bz_\GNN$ is the molecular representation obtained from a $\mathrm{READOUT}$ function using the last layer's embedding $\bh_v^K$. We follow GINE and instantiate $\mathrm{READOUT}$  as mean pooling.

\subsection*{Spectra prediction} 
Given a molecular encoding $\bz$,  we predict the  values of peak intensity at binned m/z ratios. As we discretize our data, we assume $P$ discreet bins, where each bin spans  a 1-Da range of m/z values between 0 and 1000 Da. The predicted spectrum, $\mathbf{\hat{y}}$, and mass spectra prediction loss, $\mathcal{L}$, are  defined as:
\begin{gather}
    \mathbf{\hat{y}} = \mathrm{PRED}(\bz)\\
    \mathcal{L} =  - cos(\mathbf{\hat{y}}, \mathbf{y})
\end{gather}
where $\mathrm{PRED}(\cdot)$ is neural network that predicts spectra from molecular representation $\bz$, and $cos(\cdot,\cdot)$ is the cosine similarity between the predicted spectra, $\mathbf{\hat{y}}$, and the query spectra, $\mathbf{y}$. For all $\mathrm{ PRED}(\cdot)$ functions, we  applied a two-layer MLP  augmented with bidirectional prediction mode \citep{wei2019rapid}, which increases the prediction accuracy on the larger fragments that arise due to neutral losses. 

\subsubsection*{Modeling peak dependencies using attention}
Spectral peaks may co-occur in groups reflecting a particular combination of fragments. 
Encoding such dependencies among co-occurring peaks is beneficial.
With this observation, we enhance MLP and GNN models to \textit{learn} dependencies among  peaks using two techniques, attention mechanism and  multi-task learning of predicting LDA topic distributions.

To capture dependencies among peaks, we use a multi-head  co-occurrence matrix $Q$ of $L$ heads to capture peak-to-peak co-occurrence. The spectral prediction after the attention layer, $\mathbf{\hat{y}}_{co}$, is therefore defined as: 
\begin{gather}
    \mathbf{\hat{y}}_{co} = \sum_l^L\mathbf{\hat{y}} \mathbf{Q}_l \mathbf{\tau}_l, \quad l \in \{1, ..., L\} 
 \end{gather}
where  $\mathbf{Q}_{l}$ is the $l$-th head co-occurrence matrix, sized at ${P\times P}$, where $P$ is the length of a spectrum vector. $\mathbf{\tau}_l$ is a likelihood assignments for $l^{th}$ layer, and $\sum_l^L\mathbf{\tau}_l=1$.  Both $\mathbf{Q}_{l}$ and $\mathbf{\tau}_l$ are learned during training. 

As calculating a co-occurrence matrix for all pairwise peaks ($P \times P$ dimension) is computationally expensive, we instead approximate the co-occurrence matrix by calculating a lower dimension matrix $\mathbf{D}_l$. $\mathbf{D}_{l}$ has dimensions of $P \times M$, where $M$ is the lower dimension and $M<P$ (Supplementary Fig. S7): 
\begin{gather}
    \mathbf{Q}_{l} = \mathbf{D}_{l} \mathbf{D}_{l}^\top 
 \end{gather}
 
We combine the spectra prediction (after the attention layer) with the original prediction (prior to the attention layer) as follows:
\begin{gather}
     \mathbf{\hat{y}_{update}} = \theta \mathbf{\hat{y}} + (1-\theta)\mathbf{\hat{y}}_{co}
 \end{gather}
where $\mathbf{\hat{y}_{update}}$ is the updated spectra and $\theta$ is a hyperparameter. For the rest of the manuscript, we drop the "updated" notation, and use $\mathbf{\hat{y}}$ for the updated predicted spectra. 

\subsubsection*{Predicting LDA topic distributions as a secondary task}
To further exploit peak dependencies, we run Latent Dirichlet allocation (LDA) in sklearn to learn spectral motifs. LDA assigns each spectra one or more spectral topics. We use a spectra matrix $\mathbf{Y}=[\mathbf{y}_1^\top, \mathbf{y}_2^\top, \ldots, \mathbf{y}_N^\top]$, where $N$ is the number of training data points). The LDA  model is learned based on $\mathbf{Y}$ and it assigns likelihood of topic distributions per spectra as follows:
\begin{gather}
    \mathbf{R} = \mathrm{LDA}(\mathbf{Y})
\end{gather}
where $\mathrm{LDA}(\cdot)$ is the  learned LDA model, and $\mathbf{R}$
represents the vertical stacking of topic distributions for each spectra, $\mathbf{r}$. 

We then predict the distribution of LDA topics for each spectra as an auxiliary task using Multi-task Learning (MTL). We assume the predicted topic distribution from the LDA model as ground truth labels. Learning this distribution improves molecular representation and therefore results in enhanced spectral prediction. Assuming T LDA topics, and topic distribution $\mathbf{r}$, we define the prediction function and loss for the auxiliary task as follow:
\begin{gather}
    \mathbf{\hat{r}} = \mathrm{AUX}(\bz)\\
    \mathcal{L}_{\mathrm{AUX}} =  -\sum_{t}^{T}\mathbf{r}_{t} \log(\mathbf{\hat{r}}_{t})
\end{gather}
where $\mathbf{r_t}$ is an entry for  topic $t$ in $\mathbf{r}$, $\mathbf{\hat{r}}_{t}$ is an entry for topic $t$ in the predicted spectral topic distribution  $\mathbf{\hat{r}}$, and the auxiliary function  $\mathrm{AUX} (\cdot)$ is implemented using two fully connected layers. 

\subsection*{Candidate ranking based on spectra predictions} 
Spectral prediction is performed using both the MLP-PD and the GNN-PD models on   molecules within a candidate set $C$.
Given a molecule $t$, and its spectra, $\mathbf{y}_t$, the spectral  prediction loss for $t$ and the candidates in $C$ is computed by comparing the spectral prediction against $\mathbf{y}_t$ using cosine similarity:
\begin{gather}
    \mathcal{L}_t =  - cos(\mathbf{\hat{y}_t}, \mathbf{y}_t)\\
    \mathcal{L}_c =  - cos(\mathbf{\hat{y}_t}, \mathbf{y}_c),\quad  c\in C 
\end{gather}
Based on the sorted losses, we compute 
the MLP and GNN-based rankings,  $Rank^{\mathrm{MLP}}_t$ and $Rank^{\mathrm{GNN}}_t$, respectively, corresponding to the rank of $t$ among the candidates. 


\subsection*{Rank-based ensemble model}

To use the MLP- and GNN-based spectra predictions  to maximize candidate ranking, we train an ensemble  model to learn a weighted sum of the MLP-PD and GNN-PD spectral predictions. 
As our  goal is to perform candidate ranking, the model learns the weighted sum of the two predictions based on ranking results. 
 Several steps are required for training  the ensemble scoring model.
 
First, we group  spectra in the training set based on their associated molecular formulae. 
Each such group provides a molecular candidate set (same molecular formulae, but different molecular arrangements) without the need for retrieving  candidates or spectra beyond those available in the training set.
Each spectra is treated as query spectra with a known target molecule. All other molecules in the group are considered the candidate set.

Second, for each training spectra, its known target molecule, and its candidate set, $Rank^{\mathrm{MLP}}$ and $Rank^{\mathrm{GNN}}$ are computed.
Based on the ranking of the two models, we define  labels for the classifier, $\mathbf{d}^{\mathrm{MLP}}$ and $\mathbf{d}^{\mathrm{GNN}}$, for each query spectra as follows:
\begin{gather}
   \mathbf{d}^{MLP}=\begin{cases}
    1,& Rank^{\mathrm{MLP}} \leq Rank^{\mathrm{GNN}} \\
    0,& \text{otherwise}
    \end{cases} \\
    \mathbf{d}^{\mathrm{GNN}}= 1-\mathbf{d}^{\mathrm{MLP}}
\end{gather}
A $\mathbf{d}^{\mathrm{MLP}}$ true value indicates that the query spectra is best ranked by the MLP-based model.
As the training of the ensemble model might benefit differently from each training example, we compute a weighting, $\gamma_i$, for each sample point in our training data.
A sample is more important if  the ranking of one of the models is better than the other model, indicating that the winning model has better discriminative power towards the sample. The difference in ranks between the two models also indicates better discriminative abilities. $\gamma$ is therefore computed based on the symmetric mean absolute percentage error (SMAPE) metric as follows:
\begin{gather}
      \gamma = \frac{|Rank^{\mathrm{GNN}} - Rank^{\mathrm{MLP}}|}{Rank^{\mathrm{GNN}} + Rank^{\mathrm{MLP}}}
\end{gather}
Once we obtain the weighting for each training sample, we can then train an ensemble scoring model to predict the MLP and GNN labels,  $\mathbf{\hat{d}}^{\mathrm{MLP}}$ and  $\mathbf{\hat{d}}^{\mathrm{GNN}}$, under loss $\mathcal{L}_{\mathrm{ENS}}$:
\begin{gather}
      \mathbf{\hat{d}}^{\mathrm{MLP}} = f_{\mathrm{ENS}}(\mathbf{y})\\
      \mathbf{\hat{d}}^{\mathrm{GNN}}= 1-\mathbf{\hat{d}}^{\mathrm{MLP}}\\
    \mathcal{L}_{\mathrm{ENS}} = \sum_i^{N}\gamma_i \cdot \mathrm{BCE}(\mathbf{\hat{d}}^{\mathrm{MLP}}_i, \mathbf{d}^{\mathrm{MLP}}_i)
\end{gather}
where $f_{\mathrm{ENS}}$ is a function that approximates the MLP label, and $\mathbf{y}$ is the spectra in training set.  
$\mathrm{ BCE}$ is the Binary Cross Entropy (BCE) function that compares each of the predicted probabilities to its class label, and 
$N$ is the number of training spectra. 

When using the ensemble model during the test phase, the MLP-PD and GNN-PD models predict the spectra for the candidate set. Further, $\mathbf{\hat{d}}^{\mathrm{MLP}}$ and  $\mathbf{\hat{d}}^{\mathrm{GNN}}$ are computed as above on the query spectra $\mathbf{y}$. The predicted spectra is therefore computed as:
\begin{gather}
      \mathbf{\hat{y}} = \mathbf{\hat{d}}^{\mathrm{MLP}} \mathbf{\hat{y}}_{\mathrm{MLP}} + \mathbf{\hat{d}}^{\mathrm{GNN}} \mathbf{\hat{y}}_{\mathrm{GNN}}
\end{gather}
Candidate ranking then proceeds to identify the best ranking for the query spectra by calculating the similarity on spectra predictions (discussed in \textbf{Candidate ranking based spectra predictions}).

%% file: 5_discussion.tex
\section*{Discussion}
This work proposed a novel molecule-to-spectra prediction model, ESP, for metabolite annotation. The model learns a weighting on the outputs of MLP and GNN  predictors to generate a spectral prediction for a candidate molecule. The average rank performance of the MLP and GNN models are enhanced by peak dependency considerations, including the addition of multi-head attention mechanism on peaks within the spectra and multi-tasking on spectral topics. The ensemble model significantly improves average candidate ranking performance over  MLP and GNN baseline models. 

ESP innovates over prior molecule-to-spectrum prediction approaches and provides several insights.  
First, because  our experiments highlight the merits of using both learned and fixed representations, ESP is designed to exploit both MLP and GNN models.  Fixed representations proved powerful when ranking on the most similar candidate sets, while learned representations outperformed fixed representations in all other cases. 
Second, ESP improves on spectral prediction by capturing peak co-dependencies.  ESP demonstrated two  successful yet complementary strategies: multi-head attention mechanism on  the peaks, and learning spectra motifs as secondary task via multi-task learning.  
Third, our analysis provided the insight that spectral loss weakly correlates with rank and we  showed that  spectral loss is less competitive when compared to ranking loss. Deep learning models thus will benefit from rank-learning tasks instead of relying on spectral loss, a common practice.  
Fourth,  while we performed model selection during training on the basis of average rank, we could alternatively select the model with the best top rank, or rank@k for a specified \textit{k}. Focusing on average rank allowed us to research the presence and rationale for  "difficult-to-rank" molecules.
Finally, ESP stratifies the training data by molecular formulas to create candidate sets. This strategy  proved successful in training ESP on a ranking task.  The broader idea, however, is that stratifying  training data can maximize the challenging task of within-strata discriminative learning. 

ESP utilizes additional information in the form of spectral topics to learn peak dependencies. 
Prior works have  augmented candidates with additional information to enhance  ranking predictions.  MetFrag  utilizes additional information such as citations to prioritize the candidates \citep{ruttkies2016metfrag}. CSI:FingerID utilizes fragmentation trees that best explain the spectra \citep{rauf2013finding, duhrkop2015fragmentation} to improve mapping candidates to their corresponding fingerprints \citep{vaniya2015using, duhrkop2015searching}. In addition to learning to rank, we expect auxiliary information in various forms, such as biochemical \citep{shen2019metabolic_metDNA} data or data augmentation \citep{shrivastava2021massgenie} to improve annotation. 

Another important aspect of this work is demonstrating the strong dependence of the results on the candidate set size, and more importantly, on the candidate similarity to the target molecule. Thus devising strategies to engineer an appropriate candidate set \citep{hassanpour2020biological} or to constraint de novo molecular generation becomes an important aspect of improving annotation. 
We further demonstrated that annotation results are a function of the training/test data splits, and one must consider the candidate set size and candidate similarity to the target molecule. These dependencies were consistent for ESP, and the MLP and GNN-based models. As prior works have not demonstrated otherwise, we  suspect that other annotation tools are also sensitive to data split and candidate selection.   The lack of standardized benchmark sets  and transparency in selecting test and candidate molecules prevents meaningful performance comparisons regrading the underlying computational approaches. 
We suspect that the emergence of recent deep-learning annotation models will unify our community efforts in benchmark developments.  Another important aspect of this work is showing that neural networks can be adapted to learn spectral prediction under different collision energies and adducts. However,  as demonstrated, there is a drop in performance when training/evaluating for a wide range of instrument settings. Deep learning techniques such as domain generalization and zero-shot learning stand to further enhance this aspect of metabolite annotation.

%% file: 6_SI.tex
\newpage
\section*{Supplementary Information}
\subsection*{Conceptual framework for solving  the spectrum-to-molecule approach}
A conceptual framework for solving the three subproblems involved in the spectrum-to-molecule  annotation problem (Supplementary Fig. \ref{fig:si_fromspectra}).  
\input{SI_Fig_fromspectra}

\subsection*{The Presence of Difficult-to-Rank Molecules}
We plot the distribution of the molecules at each rank (Supplementary Fig. \ref{fig:si_tail}). 
All models do well in predicting the target molecule correctly for a great number of molecules, and thus the high number of molecules at small rank values.
However, all models are challenged by  difficult-to-rank molecules that result in high rank. These molecules directly impact the average rank.

Our work herein in terms of peak dependency considerations and the learning on rank address this challenge.
Both MLP and GNN models improve in this regard when including peak dependencies 
(Supplementary Fig. \ref{fig:si_tail}A,B).  
The performance of ESP on the difficult-to-rank molecules is also improved when compared to the ESP-SL and the ESP-RU models
(Supplementary Fig. \ref{fig:si_tail}C,D), thus supporting the improvement in average rank for ESP over these two models.

\subsection*{Candidate Set Distributions}
Distributions on the  number of candidate sets retrieved from PubChem for each molecule in our test set show a long tail distribution
(Supplementary Fig. \ref{fig:si_cand}A).
The average number of candidates was 4,728 with as most as 48,292 candidates.
The similarity of the candidate sets to the target molecule  show a normal distribution
(Supplementary Fig. \ref{fig:si_cand}B).
The sets with high molecular similarities indicate that there are candidates that are difficult to rank.
Our experiments show that ESP (and other models) are more challenged by high similarity candidate sets. 
There is weak correlation between the  similarity and size of the candidate sets ($R^2$ is -0.63)
(Supplementary Fig. \ref{fig:si_cand}C).

\subsection*{MLP-PD and GNN-PD Performance as a Function of Candidate Sets}
Both MLP-PD and GNN-PD models are more challenged with increasing dataset sizes from 50, to 100 to 250 candidates
(Supplementary Fig. \ref{fig:si_mix1}). 
ESP outperforms GNN-PD and MLP-PD on the most similar candidate sets 
(Supplementary Fig. \ref{fig:si_mix2}A). 
However, on the least similar candidate set, MLP-PD outperforms ESP on ranks 1-3, and ESP outperforms MLP-PD on higher ranks (Supplementary Fig. \ref{fig:si_mix2}B).

\subsection*{MLP-PD and GNN-PD Performance on Realistic Data Splits and Full Positive Mode}
The t-SNE plot shows distinct clusters on the test molecules
(Supplementary Fig. \ref{fig:si_mix3}A). 
Under the realistic split, the models are trained on the larger clusters and tested on the smaller clusters. 
ESP marginally outperforms GNN-PD under the realistic split
(Supplementary Fig. \ref{fig:si_mix3}B). 
When training on the full positive mode dataset, performance drops for all models 
(Table \ref{Tab:S01FullMode}, 
Supplementary Fig. \ref{fig:si_mix3}C). 

\input{SI_Fig_tail}
\input{SI_Fig_cand}
\input{SI_Fig_mix1}

\input{SI_Fig_mix2}

\input{SI_Fig_mix3}

\input{SI_table1}
\input{SI_Fig_chem}

\subsection*{Annotation Example}
We provide an annotation example to highlight the influence of candidate similarity on candidate ranking performances across all models. (Supplementary Fig. \ref{fig:si_chem}).
While the three baseline models rank the target molecule among the top 3 candidates, ESP provides the correct ranking for the target molecule.  MLP-PD and GNN-PD provide  improved ranking  over the baselines, but rank the target molecule in second position.

\subsection*{Peak Dependency Modeling}
An attention layer allows modeling peak dependencies
(Supplementary Fig. \ref{fig:si_atten}). 
The spectra with attention, $\mathbf{\hat{y}}_{co}$, is computed based on $L$ attention layers. Attention is learned through a lower dimensional matrix D. 
\input{SI_Fig_atten}

\subsection*{Model Tuning}
For MLP and GNN baseline models, we follow the author's recommended guidelines on hyperparameter tuning. Otherwise, the range of hyperparameter search is specified as follows. The dimension of the two hidden layers of encoding ECFP fingerprint and instrument setting features are selected from \{64, 128, 256, 512, 1024\}. For all MLP, GNN, and ensemble classifier  training, we optimize our models with the Adam optimizer \citep{kingma2014adam} with learning rates selected among $\{10^{-2}, 5*10^{-3}, 10^{-3}, 5*10^{-4}, 10^{-4}, 5*10^{-5},\}$. We apply dropout at a rate selected from $\{0.0, 0.3, 0.5, 0.7\}$ and L2 norm at a weight selected from $\{10^{-2}, 10^{-3}, \ldots, 10^{-6}\}$. For ensemble weighting strategy, we allow a maximum of 200 epochs. 

%% file: SI_Fig_fromspectra.tex
\begin{figure}[h]
\centering
\includegraphics[width=.85\linewidth]{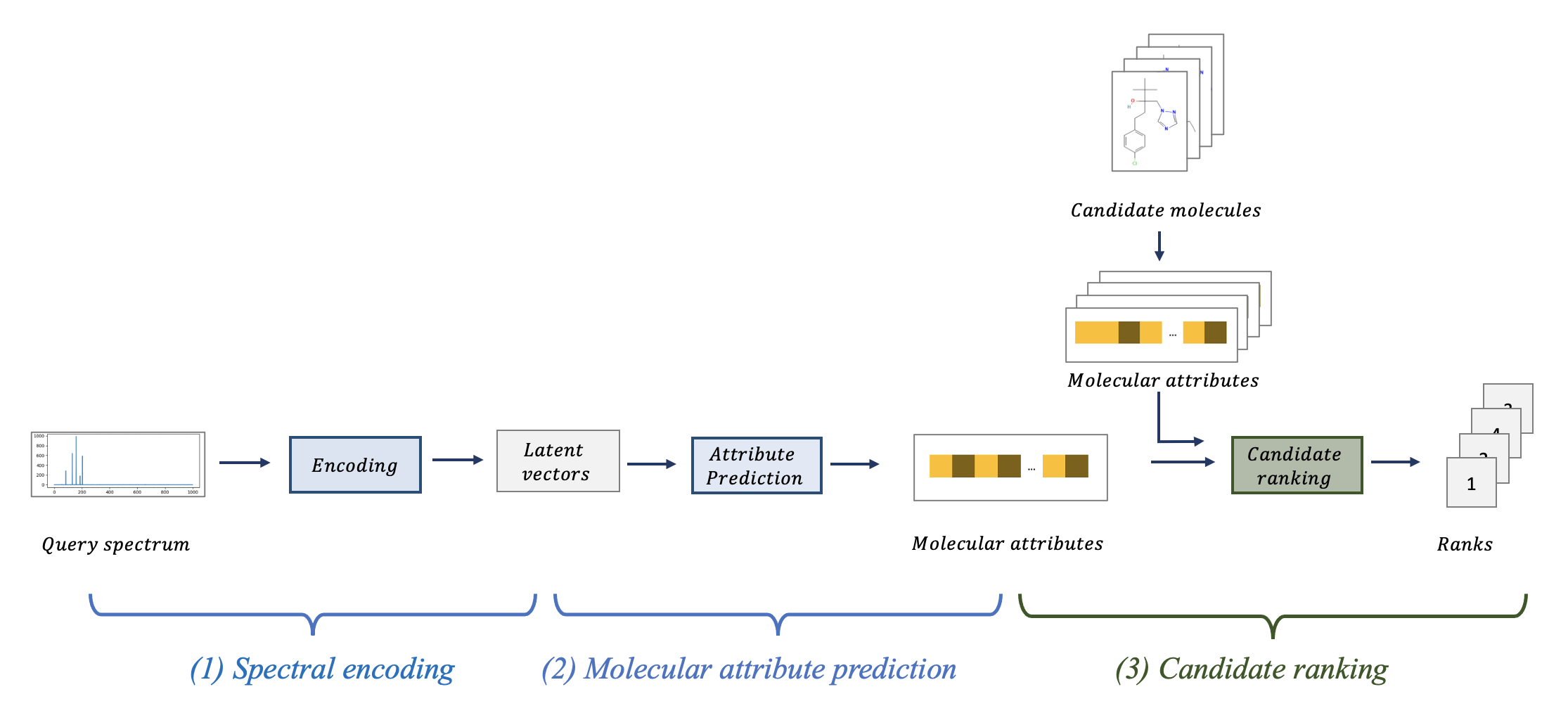}
 \caption{A conceptual framework for solving the three subproblems involved in the spectrum-to-molecule  annotation problem: (1) representation learning of query spectra, (2)  molecular attribute  prediction from  spectral representation, and (3) 
 ranking of candidate molecular attributes against predicted attributes. Shown is a prediction of molecular attributes; however, the ranking is still applicable when  predicting candidate de novo molecular structures. }
\label{fig:si_fromspectra}
\end{figure}

%% file: SI_Fig_tail.tex
\begin{figure*}
\centering
\includegraphics[width=.75\linewidth]{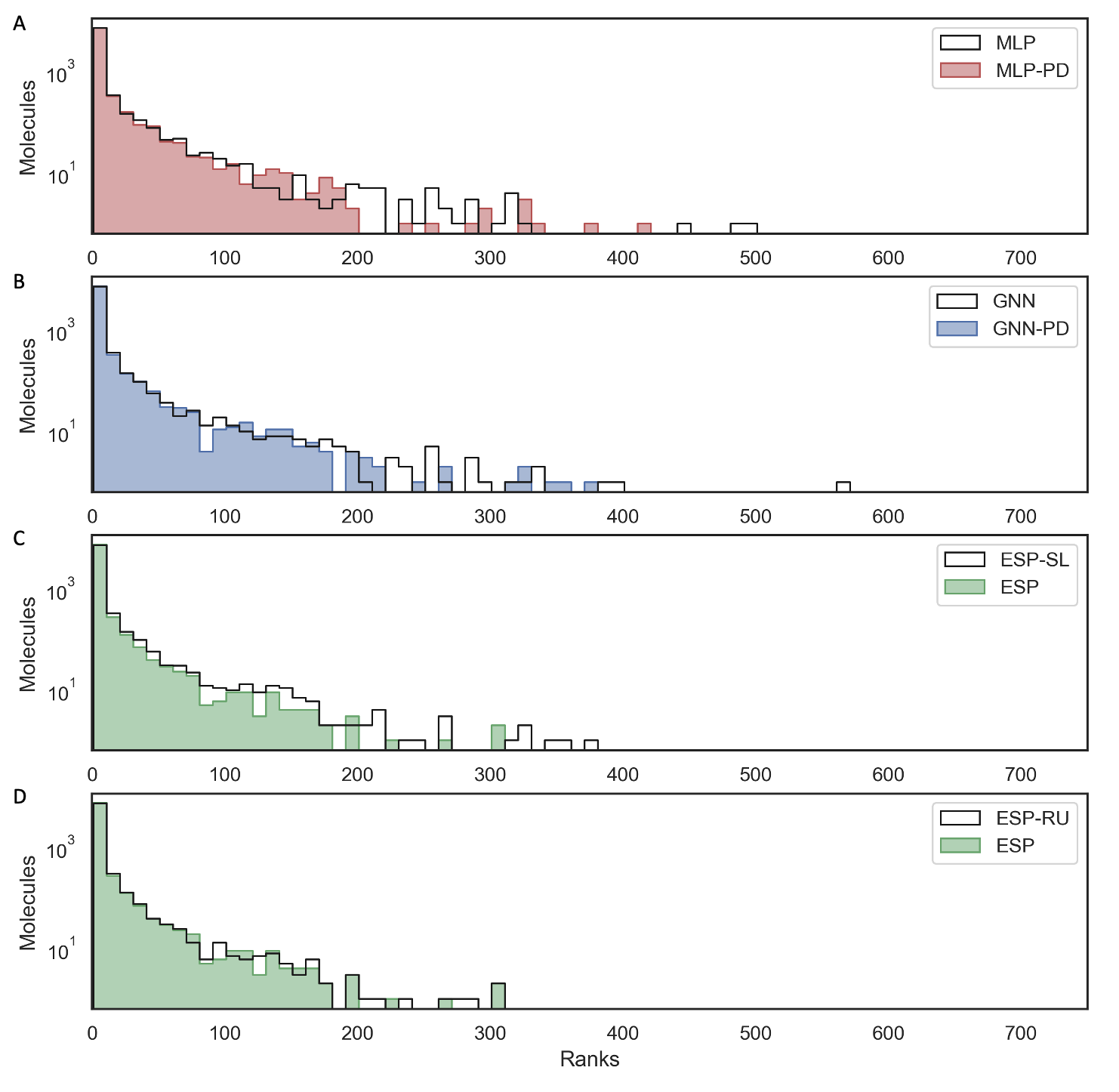}
 \caption{Number of test molecules at a particular rank. Our model improvements address the difficult-to-rank molecules, and hence improve the average rank.
A) MLP-PD  improves over the MLP model.
B) GNN-PD  improves over the GNN model.
C) ESP   improves over the ESP-SL model, which is trained  on spectral loss and not rank.
D) ESP   improves over the ESP-RU model, which weights the training examples uniformly and not by rank position and differences.
}
\label{fig:si_tail}
\end{figure*}

%% file: SI_Fig_cand.tex
\begin{figure*}
\centering
\includegraphics[width=.95\linewidth]{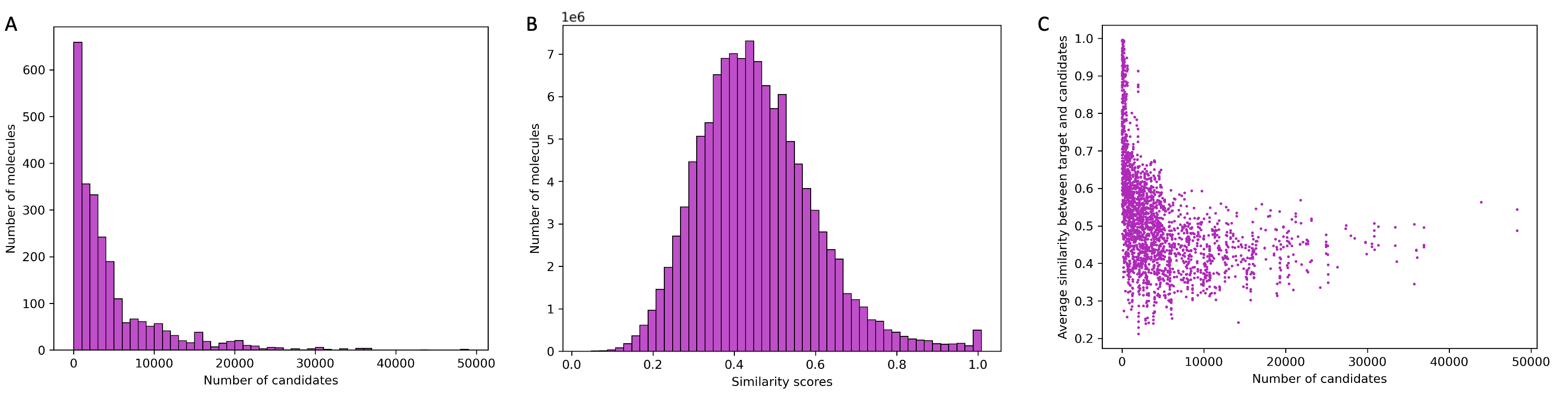}
 \caption{Profiling candidate molecules retrieved from PubChem. A) Histogram of  number of candidates  (x-axis) for the test molecules. B) Histogram of pairwise MACCS fingerprint similarity between target molecules and their respective candidates. C) Scatter plot of candidate sets showing the size of the candidate-set (x-axis) against similarity between target and candidates in each candidate set.}
\label{fig:si_cand}
\end{figure*}

%% file: SI_Fig_mix1.tex
\begin{figure*}
\centering
\includegraphics[width=.95\linewidth]{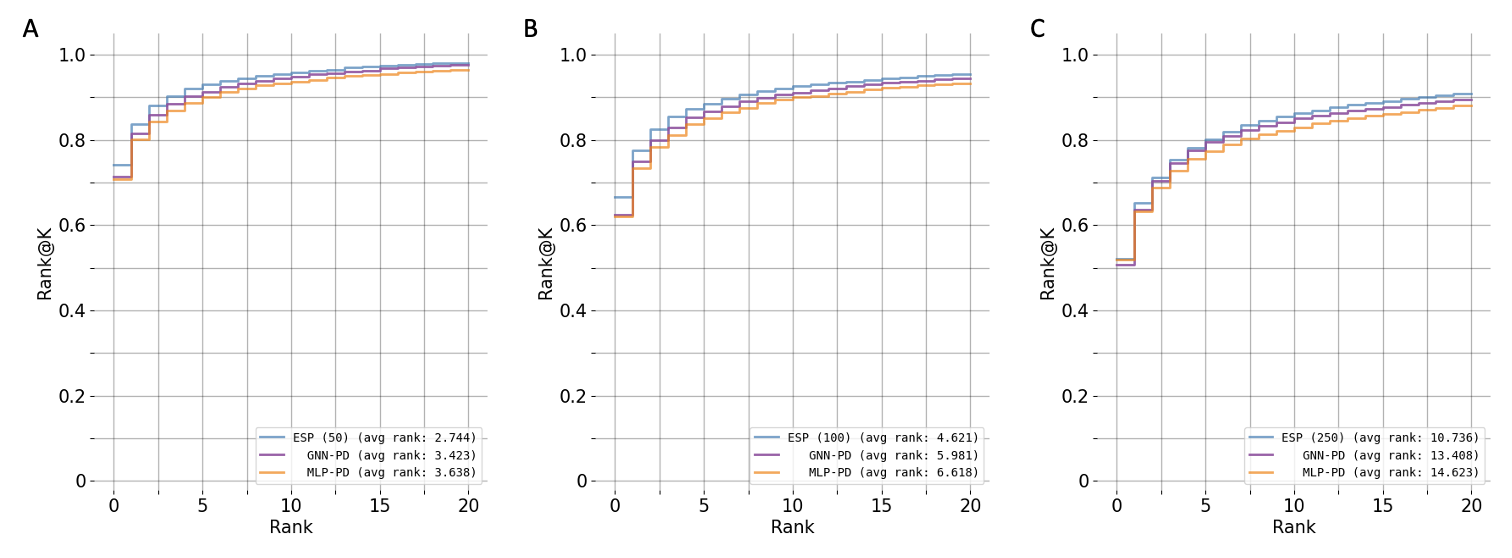}
\caption{Comparing rank@k performance for ESP, GNN-PD, MLP-PD models for different candidate set sizes of: 
 A) 50 molecules, B) 100 molecules, C) 250 molecules.}
\label{fig:si_mix1}
\end{figure*}

%% file: SI_Fig_mix2.tex
\begin{figure*}
\centering
\includegraphics[width=.65\linewidth]{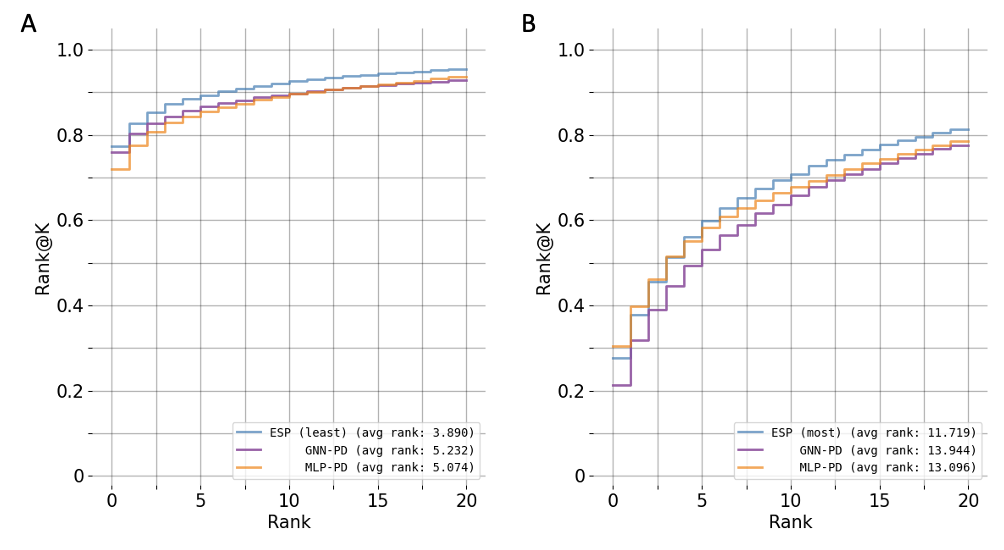}
\caption{
Comparing rank@k performance for ESP, GNN-PD, MLP-PD models for different candidate sets:
  A) least similar candidate sets, B) most similar candidate sets.}
\label{fig:si_mix2}
\end{figure*}

%% file: SI_Fig_mix3.tex
\begin{figure*}
\centering
\includegraphics[width=.95\linewidth]{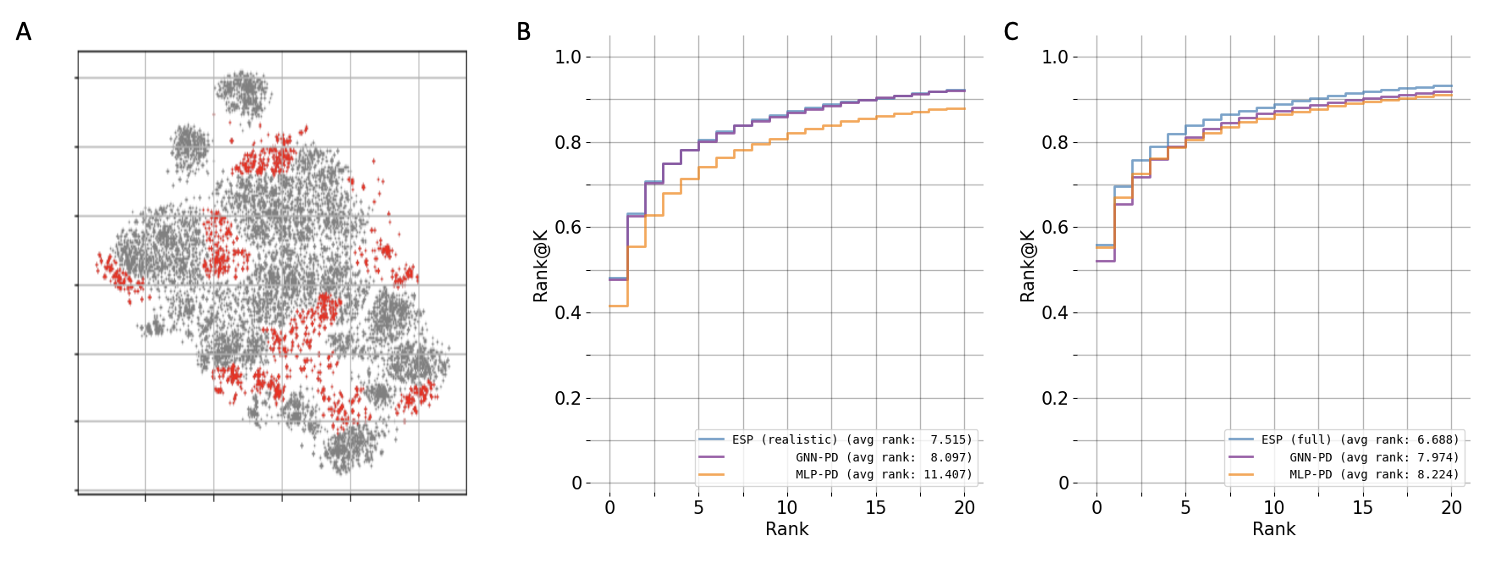}
\caption{
Analysis for realistic data splits. 
A) Realistic split t-SNE plot, where grey clusters are. used for training and red clusters are used for test).  B) Comparing rank@k performance for ESP, GNN-PD, MLP-PD models under realistic split set. C) Comparing rank@k  for the full positive ion data set.}
\label{fig:si_mix3}
\end{figure*}

%% file: SI_table1.tex
\begin{table}
\centering
\caption{Metabolite annotation evaluation on full positive ion mode.  \label{Tab:S01FullMode}} {
\begin{tabular} {ccccc}
\toprule 
&    Average rank    &    Rank@1    &    Rank@3     &    Rank@10 \\

\cmidrule(lr){2-2} \cmidrule(lr){3-5} 

& \multicolumn{1}{c}{\textcolor{blue}{The lower the better}} & \multicolumn{3}{c}{\textcolor{blue}{The higher the better}}\\
\cmidrule(lr){1-5} 
MLP	&	9.567	&	0.505	&	0.692	&	0.840\\
GNN       &	8.550	&	0.531	&	0.719	&	0.856\\

\cmidrule(lr){1-5} 

MLP-PD	    &	8.231	&	0.520	&	0.723	&	0.863\\
GNN-PD	    &	7.683	&	0.538	&	0.726	&	0.866\\
\cmidrule(lr){1-5}

ESP-SL	
                            &	8.221	&	0.553	&	0.726	&	0.854\\
\cmidrule(lr){1-5}

\textbf{ESP}
                            &	\textbf{6.688}	&	\textbf{0.558}	&	\textbf{0.757}	&	\textbf{0.881}\\
                            
\bottomrule
\end{tabular}}{}\
\end{table}



%% file: SI_Fig_chem.tex
\begin{figure*}
\centering
\includegraphics[width=.95\linewidth]{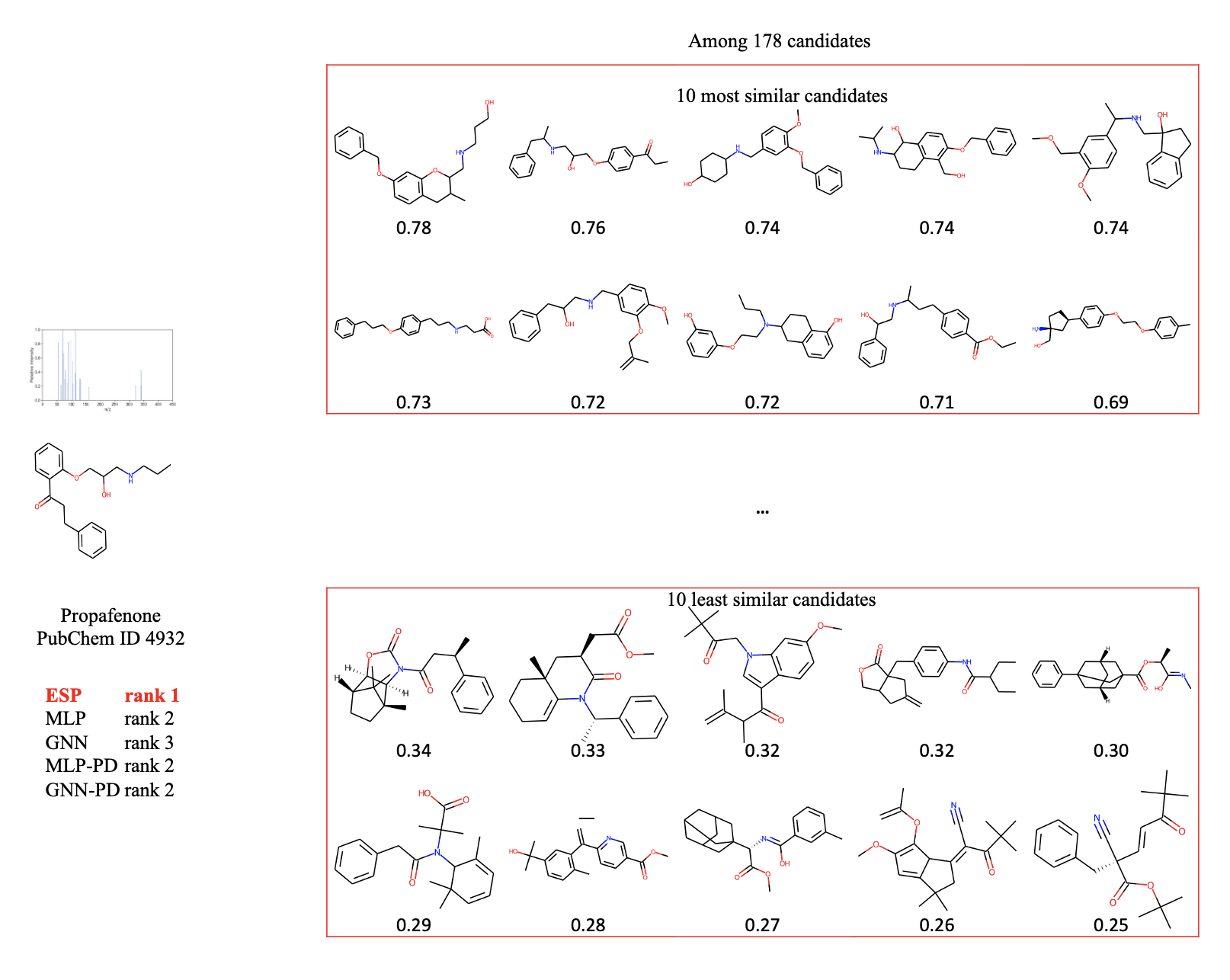}

\caption{Metabolite annotation example for target molecule  Propafenone. Shown are the 10 most and least similar candidates with their respective fingerprint similarity scores.}
\label{fig:si_chem}
\end{figure*}



%% file: SI_Fig_atten.tex
\begin{figure}[h]
\centering
\includegraphics[width=.6\linewidth]{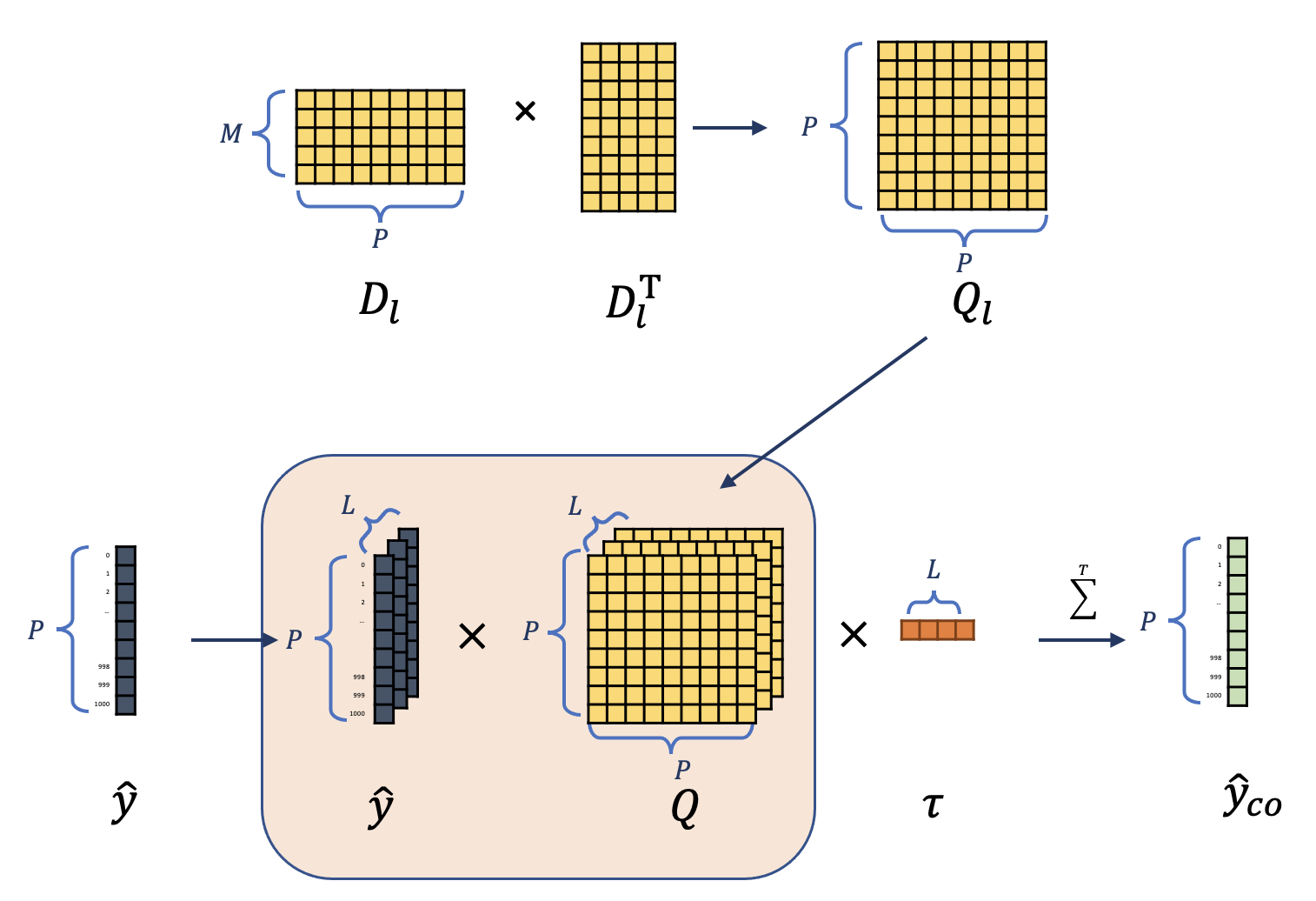}
 \caption{Using attention mechanism to capture co-occurring spectral peaks with multi-heads. The attention is applied based on previous prediction $\mathbf{\hat{y}}$,  co-occurrence matrix $\mathbf{Q}$, and weight matrix $\mathbf{\tau}$. The co-occurrence matrix, $\mathbf{Q}$, is approximated from the learned lower dimension matrix $\mathbf{D}$.}
\label{fig:si_atten}
\end{figure}